\RequirePackage{fix-cm}
\documentclass[twocolumn]{svjour3}          
\smartqed  
\usepackage{graphicx}
\usepackage{amsmath}
\usepackage{booktabs} 
\usepackage{amssymb} 
\usepackage{paralist} 
\usepackage{url}
\usepackage{xcolor}
\usepackage{placeins}

\renewcommand{\vec}[1]{\boldsymbol{#1}}
\begin{document}

\title{Anthropometric clothing measurements from 3D body scans}

\author{Song Yan \and
        Johan Wirta \and
        Joni-Kristian K{\"a}m{\"a}r{\"a}inen
}

\institute{S. Yan and J.-K. K{\"a}m{\"a}r{\"a}inen \at
  Computing Sciences, Tampere University\\
  URL: \texttt{http://www.tuni.fi} \\
  \email{First.Surname@tuni.fi}
  \and
  J. Wirta \at
  NOMO Technologies Ltd\\
  URL: \texttt{http://nomo3d.com/}\\
   \email{johan.wirta@nomo3d.com}
}

\date{Received: date / Accepted: date}

\maketitle

\begin{abstract}
  We propose a full processing pipeline to acquire anthropometric measurements from 3D measurements.
  The first stage of our pipeline is a commercial point cloud scanner.
  In the second stage, a pre-defined body model is fitted to the captured point cloud. 
  We have generated one male and one female model from the SMPL library. 
  The fitting process is based on non-rigid Iterative Closest Point (ICP) algorithm that minimizes overall energy of point distance and local stiffness energy terms. 
  In the third stage, we measure multiple circumference paths on the fitted model surface and use a non-linear regressor to provide the final estimates of anthropometric measurements. 
  We scanned 194 male and 181 female subjects and the proposed pipeline provides mean absolute errors from 2.5~mm to 16.0~mm depending on the anthropometric measurement.
\keywords{Anthropometric measurement \and 3D body model \and non-rigid ICP}
\end{abstract}
\section{Introduction}
\label{sec:intro}
\begin{figure*}[t]
  \begin{center}
    \includegraphics[width=1.0\linewidth]{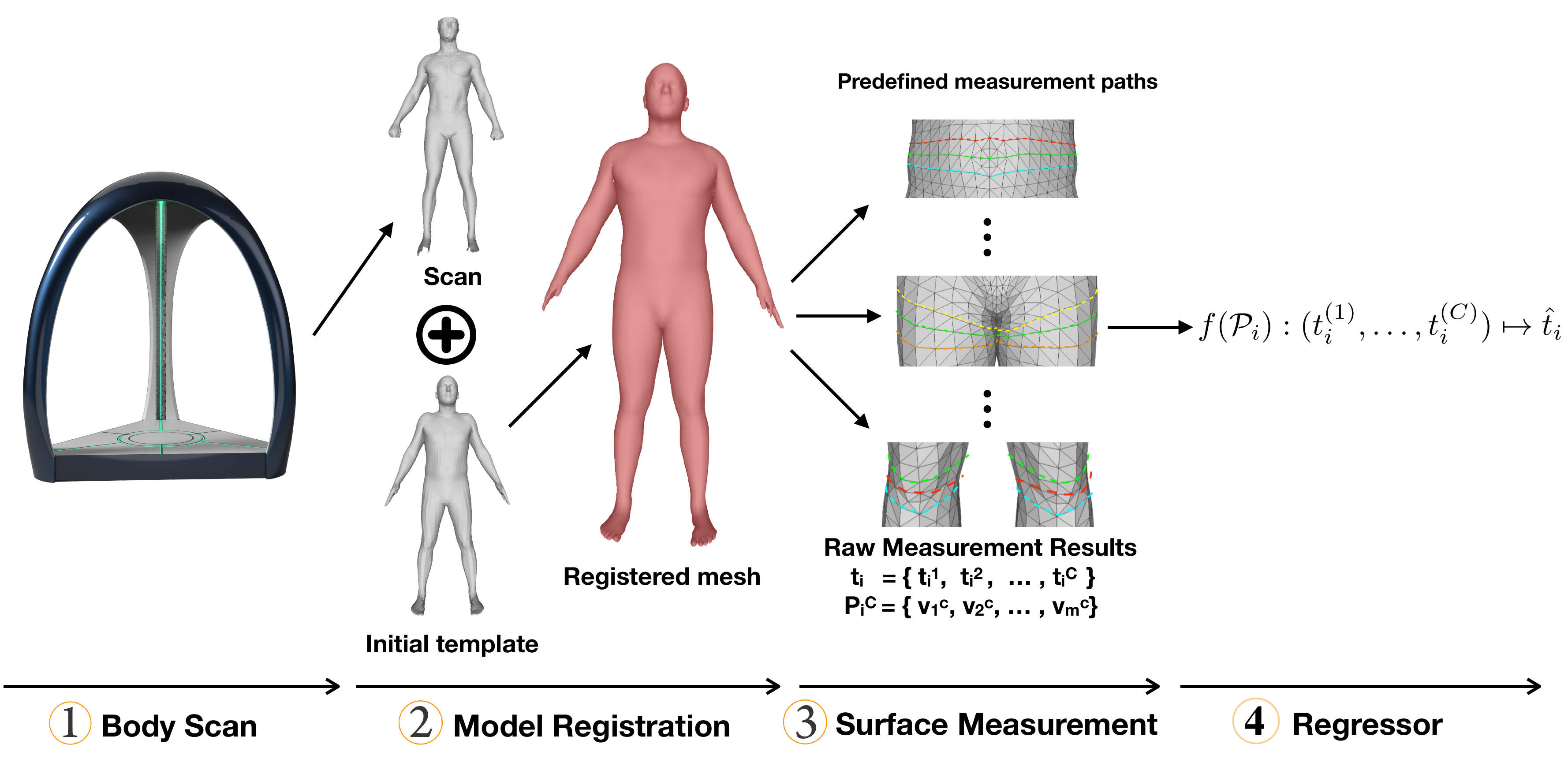}
    \caption{The proposed pipeline for measuring anthropometric clothing measurements from 3D body scans.
      A 3D point cloud is produced by a set of depth sensors (body scanner). 
      A body template is fitted (registered) to the 3D point cloud (step-2); 
      Circumference measurements are computed on the model surface (step-3); 
      Supervised regression is adopted to provide estimates of anthropometric measurements (step-4)}
    \label{fig:pipeline}
  \end{center}
\end{figure*}
Anthropometric measurements, such as chest and hip circumference or shoulder-to-shoulder distance, provide detailed information about the body shape. 
The body shape information is essential for industrial design~\cite{Park-2017-injury}, clothing design~\cite{Daanen-2008-IJCST}, medical sciences~\cite{Ogden-report} and ergonomics~\cite{Pheasant-bodyspace-book}. 
The measurements have traditionally been made manually from physical subject using a tape measure, but the raise of online shopping and personalized tools set new demand for computerized anthropometric measurements.

A standard pipeline for computerized anthropometric measurements is the following~\cite{guan2009estimating,weiss2011home,chen2011practical,baek2012parametric,wuhrer2013estimating,Tsoli-2014-wacv}
:
1) a 2D or 3D body scan producing a 3D point cloud or an initial model,
2) fitting of a pre-defined model and 3) measurements from the fitted model.
The main challenge is the step two which should provide an accurate and watertight volumetric model of a subject so that important measurements can be made on the model surface. 
Challenges arise from different sensor modalities, poses and occluded regions. 
The proposed method in this work shares the main steps of the standard pipeline (Figure~\ref{fig:pipeline}), but instead of physiologically valid model fit we adopt a non-rigid Iterative
Closest Point (ICP) registration between the model and captured point clouds.
Moreover, we do not make anthropometric measurements directly from the fitted model surface but extract a set of physiologically meaningful surface features (body circumferences) and use them to train a regressor that provides estimates of the physical anthropometric measurements.
Our main contributions are:
\begin{compactitem}
\item A full processing pipeline from 3D body scans to anthropometric measurements.
\item The body model registration step using a non-rigid ICP to fit a pre-defined model to captured body scans.
\item Non-linear regression based anthropometric measurement estimation step from circumference based intermediate features.
\item A public benchmark dataset -- NOMO3D -- with anthropometric measurement ground truth.
\end{compactitem}
Our pipeline is evaluated with the NOMO3D dataset of real male and female subjects (194 plus 181) for which we provide average accuracy and percentage of subjects whose accuracy is below the thresholds in~\cite{gordon1989anthropometric}.
\section{Related work}
\label{sec:related}

\paragraph{Anthropometric Measurement Datasets -- }
There have been several campaigns to collect 3D body scans and anthropometry ground truth for them. 
For example, the UMTRI dataset was collected to find the safest sitting posture of young children in cars~\cite{UMTRI-kids}. 
ANSUR 88 (1988) and ANSUR 2012 datasets contain 3D body scans and tape measured anthropometric measurements of US Army Force soldiers. 
ANSUR 2012 contains 4,082 male and 1,986 female subjects of varying age and 93 ground truth anthropometric measurements for each of them.
Unfortunately UMTRI and ANSUR datasets are not publicly available. 
CAESAR dataset~\cite{robinette2002civilian} is a commercial counterpart of ANSUR and contains 3D scans of 2,400 U.S. \& Canadian and 2,000 European civilians with tape measured ground truth. 
CAESAR (\url{http://store.sae.org/caesar/}) has been used in various scientific works but has not been widely adopted in benchmarking due to its price.
The main usage of UMTRI, ANSUR and CAESAR datasets is to make ``virtual tape measurements'' on the point cloud surface. In the follow-up work of the CAESAR,
Robinette and Daanen~\cite{Robinette-2006-AE} compared virtual tape measurements over two different scanners and scanning teams and showed that measurements are highly reproducible within the US Army defined error limits (cf. ANSUR experiments).
Reproducibility error in their experiments was less than $\pm 5$~mm for the most measurements. 
However, these were relative accuracies over repeated tests. 
Simmons and Istook~\cite{Simmons-2003} noted that there is substantial variation in available softwares how to measure anthropometric measurements from 3D data.
Paquette et al.~\cite{paquette2000automated} demonstrated much larger errors for 3D measurements as compared to manual tape measurements. 
They reported systematic errors of up-to $30-40$~mm despite the fact that standard measurement procedures were implemented to the softwares (ISO-8559 and U.S. Army).

\paragraph{3D Human Body Models -- }
The early works following the data campaigns above were based on ``virtual tape measurements'' where the anthropometric measurements were made manually with the help of 3D measurement software. 
If this step needs to be automated, then 3D scan data needs to be aligned with a model for which the measurement paths can be pre-defined using 3D model vertex ids. 
However, first a good 3D human body model needs to be devised. 
The model should contain intuitive parameterization for shape and pose and provide realistic body shapes. 
There are several options for scientific work. 
The most popular parametric body model is MakeHuman which is an open source project (\url{http://www.makehuman.org/}) based on an artistic body model and aiming at high quality rendering for games and movies.

However, better models are based on statistics of real human data. 
These require a single artist made initial point model which is iteratively matched to scanned point clouds in a normalized pose. 
Principal Component Analysis (PCA) over the matched model points provides natural parameterization for the shape. 
The pose can be intuitively defined by a skeleton joint model, but the final quality depends on how well the model can represent pose specific shape deformations.
One of the first attempts to create a 3D human body from PCA shape and skeleton pose is the SCAPE body model by Anguelov et al.~\cite{Anguelov-2005-siggraph}. 
Hirshberg et al.~\cite{hirshberg2012coregistration} proposed a better parametric body model for SCAPE and introduced the BlendSCAPE model. 
Other attempts are by Baek and Lee~\cite{baek2012parametric} and more recently the SMPL model by Loper et al.~\cite{SMPL:2015}. 
SMPL provides high quality models where the shape is divided to pose invariant and pose dependent deformations and the model parameters are optimized using a combination of their own dataset of $1,786$ scans and $3,800$ scans from CAESAR. 
For this work we adopt the SMPL model due to its good overall quality.

\paragraph{Computerized Anthropometric Measurements -- }
There have been several attempts to infer 3D body models from 2D RGB images.
For example, Guan et al.~\cite{guan2009estimating} proposed a method
and compared their measurements to the ground truth.
However, for many industrial and commercial applications the accuracy of 2D
measurements is insufficient. For better accuracy 3D scans are needed.

Weiss et al.~\cite{weiss2011home} propose a Kinect-based 3D body scan method that
uses the SCAPE body model. The method requires manual pose initialization and
then optimizes the model mesh using a standard ICP. Tsoli et al.~\cite{Tsoli-2014-wacv}
propose a pipeline that is similar to ours. They use the BlendSCAPE model to register
a 3D scan and then they compute various local and global features which are used
in regression. A different approach was proposed by Zuffi et al.~\cite{Zuffi-2015-cvpr}
in their ``stitched puppet'' model where the body model is divided to local templates
where ``local PCA'' matching is performed and then the local parts are globally
aligned in the next optimization step. Wuhrer et al.~\cite{wuhrer2013estimating}
introduce an inverse problem of ours where a 3D body model is estimated from
the given 1D anthropometric measurements.

The above works particularly address the problem of unknown pose. However,
we believe that a fixed pose can be assumed for many applications since
customers can be assumed co-operative.  Therefore, the process can be drastically
simplified and provide accurate results.

\section{3D body scanning}
\label{sec:scanning}
\begin{figure}[h]
  \begin{center}
    \includegraphics[width=0.8\linewidth]{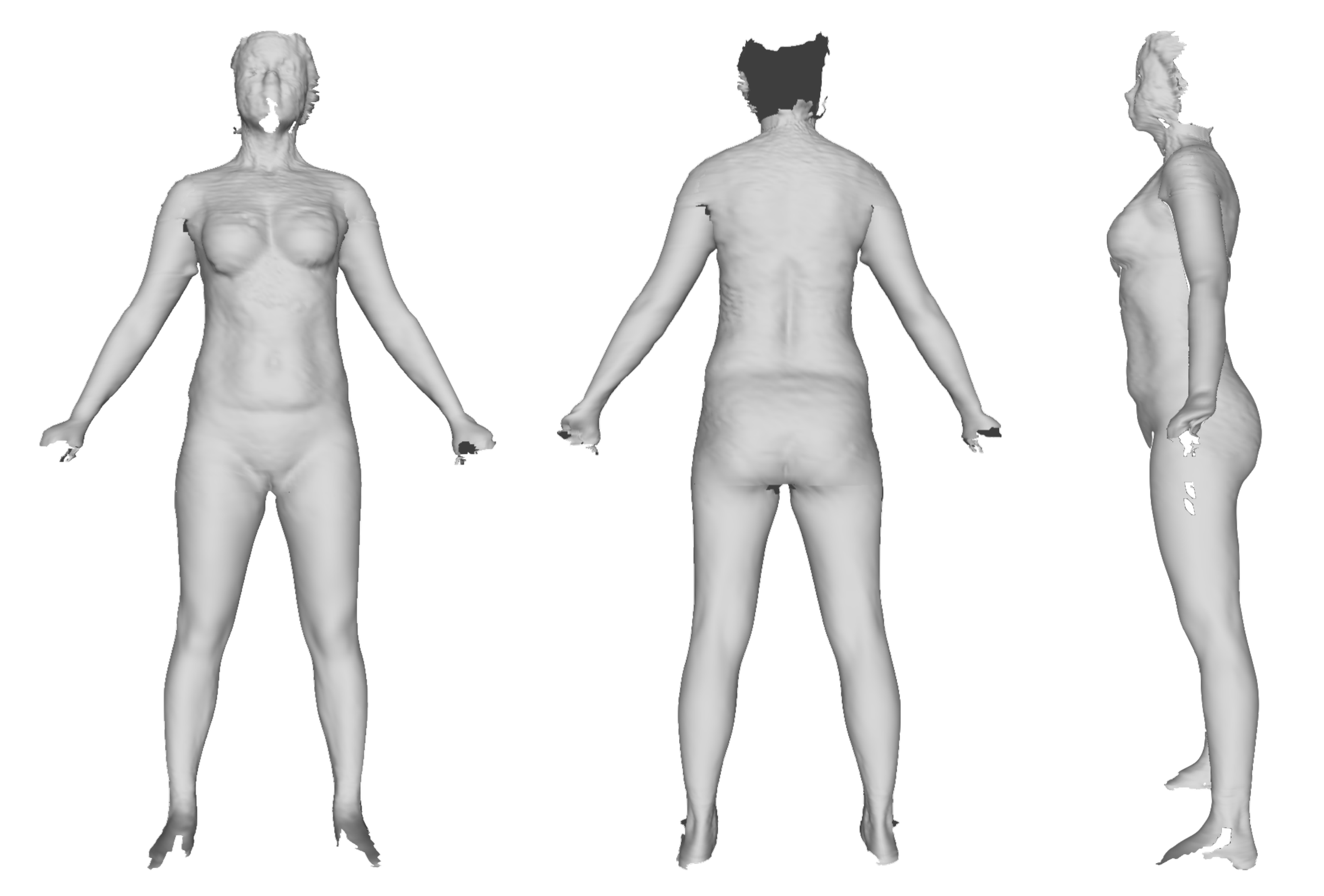}
    \caption{A 3D body scan (point cloud) captured by TC2 body scanner.
      The scanner cover most of the body surface and missing parts occur only in the
      head and feet regions}
    \label{fig:TC2_scan_sample}
  \end{center}
\end{figure}
Recently, novel single depth-sensor based body
  scanning approaches have been proposed, for example,
  Bo\-dy\-Fu\-sion~\cite{BodyFusion} and DoubleFusion~\cite{DoubleFusion},
  but since 3D scanning is out of the scope of this work, a
  commercial 3D body scanner was used.
Our dataset was collected using a commercial TC2 body scanner
(\url{https://www.tc2.com})
that uses off-the-shelf depth sensors (Intel RealSense R200).
Inside the scanner, subjects were instructed to step on the rotating
platform and take a standing pose with the feet at around their shoulder
width apart and the arms slightly raised to create a gap between the arms
and torso. The platform then rotates around once, during which three depth
sensors produce a raw 3D scan of the customer and the process
takes a few seconds (Figure~\ref{fig:TC2_scan_sample}). The test subjects wore
tight fitting underwear-like sport costumes.
The scanner outputs a triangulated mesh structure in the
regular OBJ file format. Each triangulated mesh contains on average 57,000
vertices and around 113,000 faces. For our experimental studies, we
scanned 194 men and 181 women. Scanned persons were instructed to wear
tight underwear.
\section{Model registration}
\label{sec:registration}
\subsection{SMPL body model}\label{sec:model}
\begin{figure}[h]
  \begin{center}
    \includegraphics[width=0.8\linewidth]{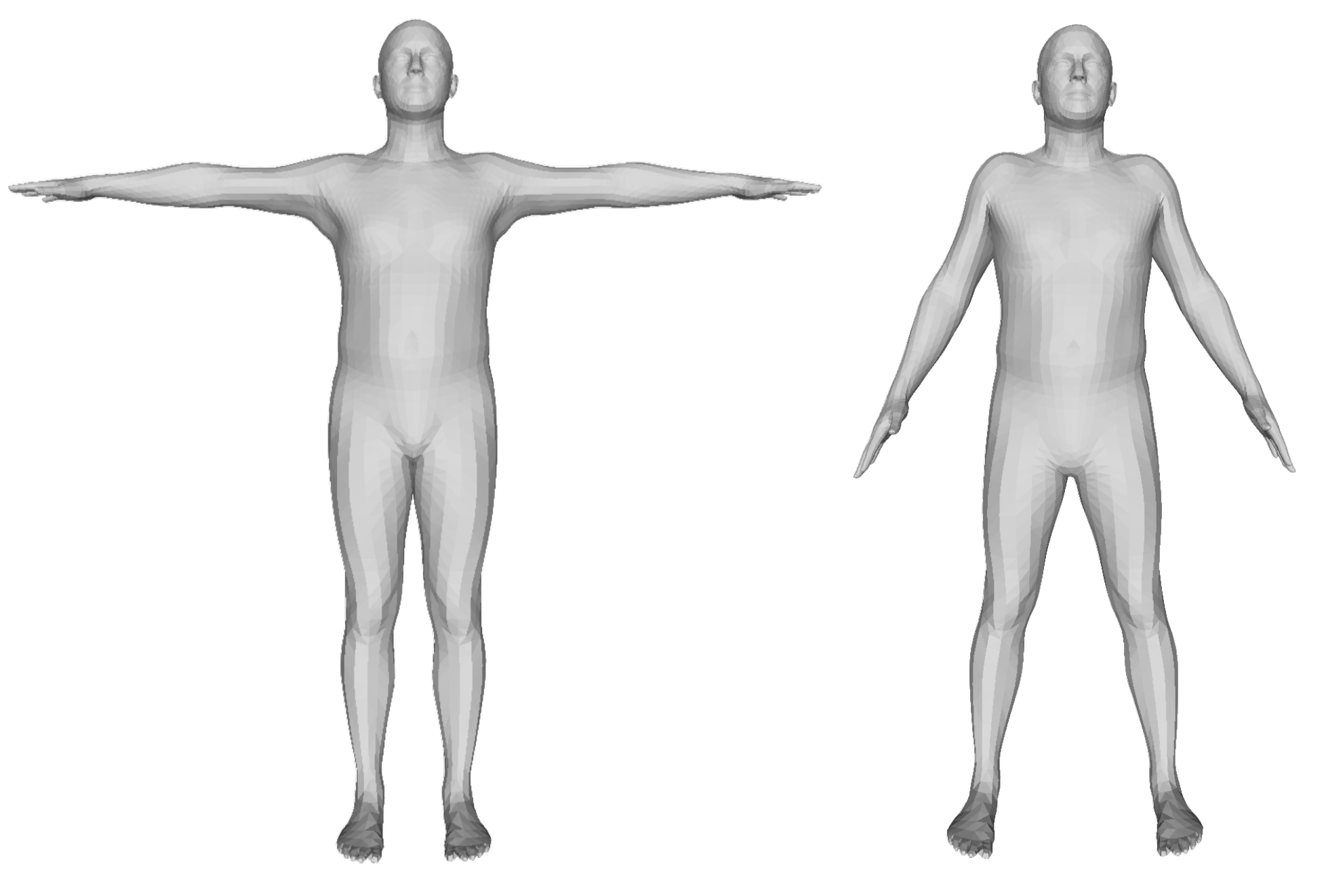}
    \caption{We adopted the Skinned Multi-person Linear Model (SMPL)~\cite{SMPL:2015}
      for our framework since it provides intuitive model parameterization and high
      quality models.
      A SMPL model in its canonical (zero-pose) position (left) and the model in the initial position
      that corresponds to the instructed pose in our body scans (right)}
    \label{fig:smpl_init_template}
  \end{center}
\end{figure}

The popular 3D human body models MakeHuman, SCAPE~\cite{Anguelov-2005-siggraph},
BlendSCAPE~\cite{hirshberg2012coregistration} and SMPL~\cite{SMPL:2015}
(see Section~\ref{sec:related} for details) share similar model parameterization
$\left\{\mathcal{T},\mathcal{S},\vec{\theta}\right\}$ where
$\mathcal{T}$ is the initial model in a ``canonical shape'' and ``canonical pose'',
$\mathcal{S}$ defines the shape deformation and $\vec{\theta}$ defines the
pose. Pose parameterization is intuitive and typically based on a skeleton rig
of $K$ skeleton joints. A pose is encoded to the 3D rotation angles
of $K$ joints in $\vec{\theta}$.
Each vertex location in $\mathcal{T}$ is relative to a specific skeleton part
or parts and therefore the whole point cloud deforms.
Parameterization of the shape is more difficult to model since parameters need to capture
shape statistics of the human population.
The standard approach is to use Principal Component Analysis (PCA) where principal
components represent the most important axes of variation in the population.
In the PCA space any
shape can be reconstructed by linearly adding $|\vec{\beta}|$ principal
directions to a mean shape $\mathcal{T}$ (the zero shape):
\begin{equation}
  \qquad \qquad \qquad \mathcal{T}+B(\vec{\beta}) = \mathcal{T}+\sum_{n=1}^{|\vec{\beta}|}\beta_n\mathcal{S}_n \enspace .
\end{equation}
Often as few as $|\vec{\beta}|=10$ principal component vectors provide sufficient
accuracy for applications where subtle details are not important. For our work we selected the 
{\em Skinned Multi-person Linear Model} (SMPL) by Loper et al.~\cite{SMPL:2015}
since it provides very competitive accuracy and the original implementation is publicly
available.

SMPL mesh model contains $N = 6,890$ vertices ($13,766$ faces) and $K = 23$ skeleton joints.
The mesh has the same topology for men and women, spatially varying resolution, a
clean quad structure, segmentation into parts, initial blend weights, and a skeletal rig.
A particular detail that makes SMPL registration more accurate than its competitors is that it
divides the shape deformation to pose independent deformation $B_S(\vec{\beta})$ and
pose specific deformation $B_P(\vec{\theta})$ which are summed to define the final shape.
Notably the shape deformation parameters are also used to predict the rotations of the
$K=23$ skeleton joints $J(\vec{\beta}): \mathbb{R}^{|\vec{\beta}|} \rightarrow \mathbb{R}^{3K}$.
We re-defined the SMPL zero-pose to correspond to the pose subjects were instructed to take (Figure~\ref{fig:smpl_init_template}).
\subsection{Non-rigid ICP registration}
\begin{figure}[h]
  \begin{center}
    \includegraphics[width=0.8\linewidth]{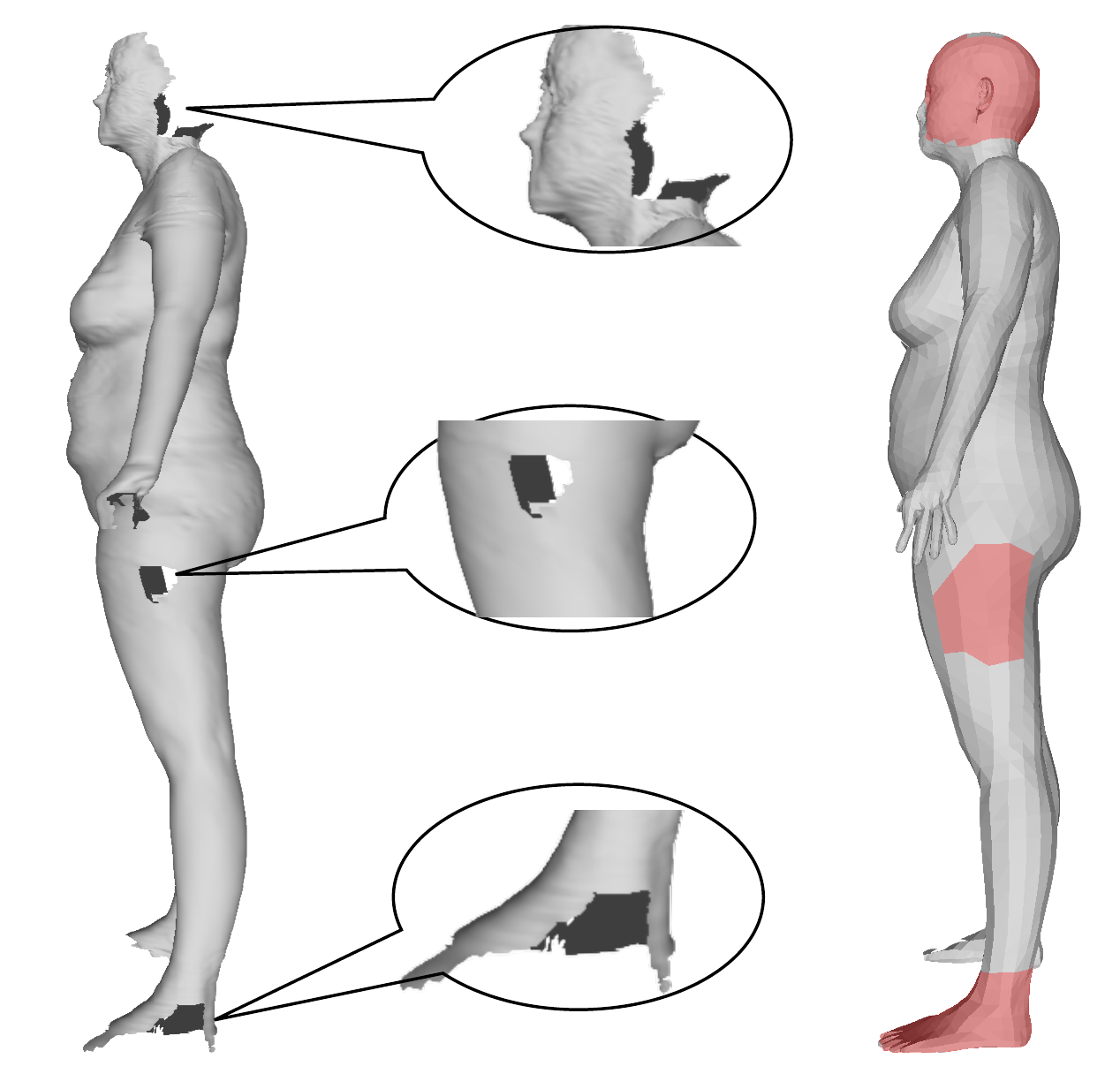}
    \caption{A scanned point cloud contains holes and measurement noise, but registration of the 3D body
      model (red) is robust to these distortions and achieves an accurate - ``skin level'' -
      registration which is essential for accurate anthropometric measurements in the next stage}
    \label{fig:registration}
  \end{center}
\end{figure}
The goal of the body model registration to the scanned point cloud is to provide
``skin level registration'' where the two surfaces, the model and the scan,
overlay almost perfectly (Figure~\ref{fig:registration}). This is a challenging task
since
a)~points contain measurement noise,
b)~large point regions may be missing and
c)~the model points do not exactly match the scan point locations. To make
the final anthropometric measurements accurate in the next processing stage
we need a registration method that is accurate and robust
to the aforementioned non-idealities.

A core component in constructing the SCAPE, BlendSCAPE and SMPL datasets is
an artistic generated point model and an algorithm to register the model
to real human scans. However, these algorithms perform complex optimization
and must be manually initialized.
Therefore the artistic models and special algorithms have not been used outside body model generation.
However, the final body models, SCAPE, BlendSCAPE and SMPL, provide intuitive parameterization
as discussed in Section~\ref{sec:model} and registration can be defined
as an optimization problem where a few pose and shape parameters
$\left\{\mathcal{S},\vec{\theta}\right\}$ are optimized to minimize a
registration error.  Skin level registration requires a large
number of PCA components for the shape and therefore we
take an alternative approach from the generic point cloud matching literature.

Several comparison of generic registration methods exist. For example,
Bogo et al.~\cite{bogo2014faust} introduced the FAUST dataset for
comparing non-rigid registration methods. In their experiments,
several popular methods, e.g., Generalized Multi-Dimensional Scaling
(GMDS)~\cite{Bronstein-2008-book}, M\"obius voting~\cite{Lipman-2009-siggraph}
and Blended Intrinsic Maps (BIM)~\cite{Kim-2011-siggraph}, did not perform well
since these methods assume that both inputs are watertight and have the
same topology. However, the baseline point cloud matching method,
Iterative Closest Point (ICP), does not require such assumptions.

There are two extensions of the baseline ICP that are suitable for
human body point clouds: Amberg et al.~\cite{amberg2007optimal} and
Schneider et al.~\cite{schneider2009fast}. Since the 3D scans often contain
holes (Figure~\ref{fig:TC2_scan_sample}) we adopted the Amberg et al.
approach that explicitly handles missing points.
The challenge is two-fold - we want to retain the global convergence properties of ICP while
still allow local deformations to the skin level. Local deformations make this ICP
non-rigid.

The starting point of our algorithm is a pre-aligned model defined by
$\left\{\mathcal{T},\beta_i,\theta_k\right\}_{i=1,\ldots,|\beta|,k=1,\ldots,K}$ that brings
the SMPL template to approximate correspondence with the obtained scan point cloud
$\mathcal{T}_{scan}$. A simple procedure for pre-alignment is described in
Section~\ref{sec:pre_alignment}. If we define the pre-aligned model as $\mathcal{V}$
then the problem is to find optimal values for the alignment parameters $\mathcal{X}$
so that $\mathcal{V}(\mathcal{X})$ registers the template points to the surface
points $\mathcal{T}_{scan}$.

To solve the optimal parameters $\mathcal{X}$ an energy function of three
terms is defined~\cite{amberg2007optimal}: 
\begin{equation}
  \qquad \qquad E(\mathcal{X}) = E_d(\mathcal{X}) + \alpha E_s(\mathcal{X}) + \beta E_l(\mathcal{X}) \enspace .
\end{equation}
$E_d$ is the standard ICP {\em distance term} between the
model and scan points
\begin{equation}
  \qquad \qquad E_d(\mathcal{X}) = \sum_{v_i \in \mathcal{V}}w_i\hbox{dist}^2\left(\mathcal{T}_{scan},\mathcal{X}_iv_i\right)
\end{equation}
where $\mathcal{X}_i$ is a linear mapping of a single model vertex $v_i$ to
correspondence in $\mathcal{T}_{scan}$. $w_i$ defines whether a model point
has a correspondence in scan ($w_i = 1$) or not ($w_i = 0$). $E_s$ is
a local {\em stiffness term}
\begin{equation}
  \qquad \qquad E_s(\mathcal{X}) = \sum_{i \in \mathcal{N}_j} \| (\mathcal{X}_i - \mathcal{X}_j) \hbox{diag}(1,1,1,\gamma)\|^2_{F}
\end{equation}
where $\| \cdot\|^2_F$ is the matrix Frobenius norm. The stiffness term enforces similar
transformations between neighbor vertices $\mathcal{N}_j$ of the model vertex $v_j$.
$\gamma$ is used to weight differences in the rotational and
skew part of the deformation against the translations part
of the deformation ($\gamma=1$ in the experiments). The third energy
term is a {\em landmark term}
\begin{equation}
  \qquad \qquad \qquad E_l(\mathcal{X}) = \sum_{\vec{v}_i,\vec{l} \in \mathcal{L}} \| \mathcal{X}_i\vec{v}_i - \vec{l}\|^2 \enspace .
\end{equation}
The landmarks $\mathcal{L}$ are pre-defined and important positions in the model and this term enforces
them to be registered accurately. The landmark term improves registration significantly, but
requires manual labeling of selected keypoints and is therefore omitted in our experiments.

The algorithm in~\cite{amberg2007optimal} uses locally affine regularization
which assigns an affine transformation to each vertex and minimizes the difference in
the transformation of neighboring vertices. The deformation parameters
$\mathcal{X}$, which would be applied on source vertices to generate the target
surface deformation, are obtained by minimizing the cost function in
Eq.~\ref{eq:nonrigid_icp_eq} directly and exactly.
\begin{flalign}
\label{eq:nonrigid_icp_eq}
\begin{split}
\qquad \qquad \bar{E}(\mathcal{X}) &= \left\lVert
	\begin{bmatrix}
		\alpha\boldsymbol{M}\otimes\boldsymbol{G}\\
		\boldsymbol{WD}\\
		\beta\boldsymbol{D}_{L}
	\end{bmatrix}
	\mathcal{X}
	-
	\begin{bmatrix}
	\boldsymbol{0}\\
	\boldsymbol{WU}\\
	\boldsymbol{U}_{L}
	\end{bmatrix}
\right\rVert^{2}_{F} \\
&= \Vert \boldsymbol{A}\mathcal{X} - \boldsymbol{B} \Vert^{2}_{F}
\end{split} \enspace .
\end{flalign}
The cost function $\bar{E}(\mathcal{X})$ takes its minimum at
$\mathcal{X} = (\boldsymbol{A}^T\boldsymbol{A}^{-1})\boldsymbol{A}^T\boldsymbol{B}$. 
In the above equation $\boldsymbol{M}$ is the node-arc incidence matrix of the
template mesh topology, and $\boldsymbol{G} := diag(1,1,1,\gamma)$ is a weighting matrix,
$\boldsymbol{W} := diag(w_1, ..., w_n)$  represents the weighting matrix in which
$w_i = 0$ if template vertices $v_i$ corresponds to missing data in the target mesh and $n$
represents the number of template vertices, $\boldsymbol{D}$ is the sparse matrix of
template vertices mapping the $4n\times3$ deformation parameters $\mathcal{X}$, $\boldsymbol{U}$
is the matrix of the correspondence points on the target mesh, $\boldsymbol{D}_L$ and $\boldsymbol{U}_L$
are the pre-defined landmarks on the template mesh and their correspondence points on the target
mesh respectively, the Kronecker product is denoted by $\otimes$. $\alpha$ and
$\beta$ are the penalty terms that balance the two corresponding energy functions with
respect to the standard ICP term $E_d$.

The whole registration process consists of two loops. 
In the outer loop a series of deformations of the template are performed for each
stiffness $\alpha^i \in \{\alpha^1, ... , \alpha^n\}$, where $\alpha^i > \alpha^{i+1}$. 
These $\alpha$ values guarantee the registration process from a global deformation
to more localized ones. In our experiments $\alpha$ values are set to
from $100$ to $1$ by step size $1$. In the inner loop a deformation
$\mathcal{X}$ for a fixed stiffness term $\alpha^i$ and preliminary correspondences is found. 
Preliminary correspondences are found by a nearest point search. 
The optimal deformation $\mathcal{X}$ is determined until
$||\mathcal{X}^j - \mathcal{X}^{j-1}|| < \epsilon$, where $\epsilon$ is the threshold.
\subsection{Pre-alignment and Initialization Procedures}\label{sec:pre_alignment}
A simple pre-alignment procedure is performed before non-rigid ICP registration. 
Generally the mis-alignment of registration is partly raised by wrong scales,
face orientations, the different center points of subjects. 
To depreciate it, firstly we scale all scans into the same unit of measurement (meter)
as the SMPL model meshes; we then rotate all scans to make sure that they face the same direction. 
Compared to the previous works which adopt the mean coordinate of vertices as the center
points and align all meshes into the same center point, we additionally align all samples
into the same lowest point (Z-axis). The center points change dramatically since the missing
parts on scans and bring negative effects on registration. A standard point $(x, y, 0)$ is
employed as the lowest point for all meshes. After the pre-alignment procedure, all
scans and the SMPL models are standing on the $X-Y$ plane and facing to $Y$-axis direction
with the same scale.

The height of the SMPL model is controlled by the first shape parameter $\beta_1$.
To obtain a suitable initial value for $\beta_1$,
we utilize a simple linear function over the heights of the training set scans
to estimate the parameter $\hat{\beta_1} \approx \beta_1$.
To initialize the pose parameters, 
we start from the pose $\vec{\theta}$ (on the right in Figure~\ref{fig:smpl_init_template})
and iteratively test a number of arm angle shifts to match with the target scan.
These initialization procedures aid convergence and improve accuracy, but their
effect is not significant.

\section{Anthropometric Measurements}
The proposed pipeline outputs estimates of the target physical anthropometric measurements from
a fitted model (Section~\ref{sec:registration}) by first calculating
{\em circumference paths} through the model points
(Section~\ref{sec:surface_measurements}) and then estimating
the physical measurements from the path distances by 
non-linear regression (Section~\ref{sec:regression}).

\subsection{Surface Measurements}\label{sec:surface_measurements}
\begin{figure}[h]
  \centering
  \includegraphics[width=0.6\linewidth]{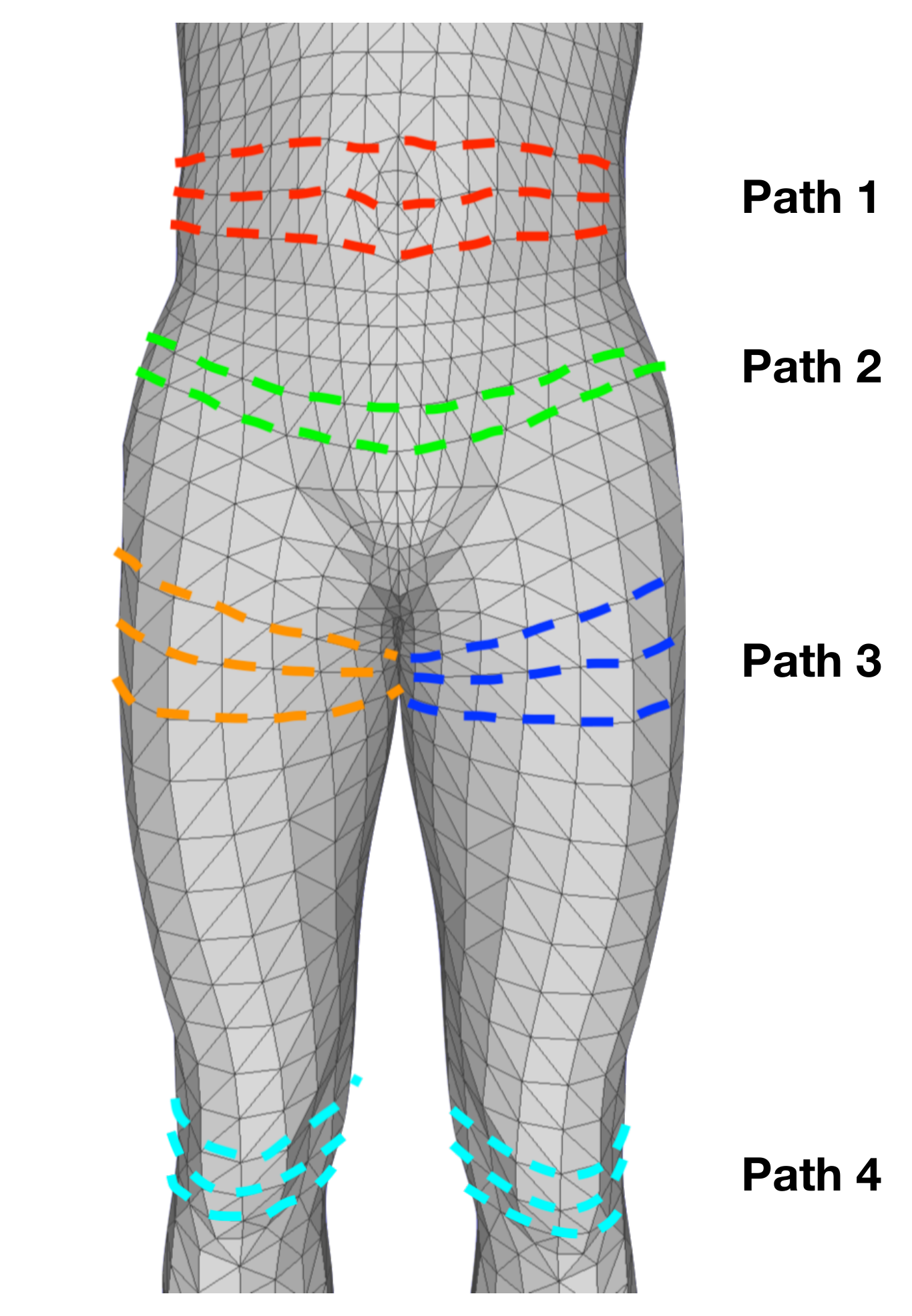}
  \caption{Distances of circumference paths through mesh vertices of a
    registered SMPL body model are used as features for regression. Multiple paths
    (dotted red, green and blue lines) are used to estimate a single
    anthropometric measurement. Example circumference paths: 
    Path 1: {\em NaturalWAIST}; Path 2: {\em Hip};
    Path 3: {\em Thigh}; Path 4: {\em Knee}}
  \label{fig:predefined_circ_path_fig}
\end{figure}

The registration process brings two main benefits: (a) it produces a hole-free mesh without
missing body parts and reduces the point cloud noise;
and (b) registered meshes of all subjects are in the same topology that facilitates
finding the corresponding vertices of the pre-defined circumference paths.

For each anthropometric measurement
$t_i$ we define a set of surface circumference paths. The path lengths
$t_i^{(1)}$, \ldots, $t_i^{(C)}$ are used as features for regression.
The paths are defined as sets of vertices in the model
$\mathcal{P}^{c}_i = \left\{\vec{v}^{c}_1, \vec{v}^{c}_2,\ldots, \vec{v}^{c}_m\right\}$.
The length of a circumference path is the sum of edge lengths through the
defined path (Figure~\ref{fig:predefined_circ_path_fig}).
The selected circumference paths were not optimal, but manually set 
near the
true anthropometric measurement locations. It was assumed that multiple paths
provide extra robustness to shape deformations (see the ablation study in the
experimental part of our work).

\subsection{Non-linear regression}\label{sec:regression}
The purpose of a suitable regressor is to find a mapping $f(\cdot )$ such that
\begin{equation}
  \qquad \qquad \qquad f(\mathcal{P}_i): (t_i^{(1)}, \ldots, t_i^{(C)}) \mapsto \hat{t}_i
\end{equation}
where $\hat{t}_i$ is the estimate of the true anthropometric clothing measurement $t_i$.
The most straightforward solution is the ordinary least squares (linear regression)
which finds a solution $\vec{\omega} = (\omega_0,\omega_1,\ldots,\omega_C)^T$ that
minimizes the squared loss over training subjects $i$
\begin{equation}
  \qquad \qquad \qquad \sum_i \left(t_i - \vec{\omega}^T\vec{t}_i\right)^2
\end{equation}
where $t_i$ is a training set the ground truth value and
$\vec{t}_i = (t_i^{(1)}, \ldots, t_i^{(C)})^T$ are the computed
circumference path distances for this specific anthropometric
measurement. Linear regression with regularization (ridge regression)
minimizes the squared loss with a weight penalty term $\lambda$
\begin{equation}
  \qquad \qquad \qquad \sum_i \left(t_i - \vec{\omega}^T\vec{t}_i\right)^2+\lambda||\vec{\omega}|| \enspace .
\end{equation}
The are also more advanced extensions of linear regression, such as
Elastic Net Regression~\cite{Zou2005_elasticnet}, and other learning
based regressors such as Support Vector Regression (SVR)~\cite{Smola2004_svr}.
We compare several popular regression methods in our ablations studies.
\section{Experiments}
\label{sec:experiments}
\subsection{Dataset and Settings}
We collected a set of 3D scans using the commercial scanner (Section~\ref{sec:scanning}).
The dataset - NOMO3D - consists of 194 male and 181 female scans. For each subject,
a clothing expert (tailor) made the actual anthropometric measurements (15 male and
19 female). All results are average performance over 5-fold cross-validation.
\paragraph{Method Evaluation -- }
We employ the Mean Absolute Error (MAE) as the error metric between the ground truth and estimated
anthropometric measurements. For each measurement $i$, Mean Absolute Error $\epsilon_i$, over all
subjects $j$ was obtained as
\begin{equation}
\label{eq:mae_eq}
\qquad \qquad \qquad \epsilon_{i} = \frac{1}{|j|}  \sum_{j=1}^{|j|} |t^{(j)}_{i} - \hat{t}^{(j)}_{i}| \enspace .
\end{equation}
In addition to the measurement specific MAEs we also computed the average MAEs
over all measures. All numbers were measured in millimeters (mm). Moreover,
for each measurement we also report the proportion of the test samples for
which the accuracy was below the defined error limits
in~\cite{gordon1989anthropometric} as {\em Success rate}.

\paragraph{Computational complexity -- }
The most time consuming part is the non-rigid ICP registration. Matlab code
was adapted from~\cite{amberg2007optimal} and it runs approximately 2 minutes
on each scan. The pre-alignment and initialization procedures are very fast,
less than a second, as well as the regression which is also computationally
fast.
\subsection{Results}
\begin{table*}[h]
      \centering
      \caption{Average 5-fold ($80\%$ for training and $20\%$ for testing)
          performance (Mean Absolute Error) and success rate (a proportion of the test
          samples within the error limits in~\cite{gordon1989anthropometric}) of
          anthropometric measurements. "C" denotes the number of circumference paths used
          in estimation. "best single" is the best single path performance.
          ''+SVR'' uses SVR regression for the estimates. 
          "L-BFGS-B + SVR" uses the SMPL model fitted by the L-BFGS-B optimizer.
        \label{tab:results1}
      }
        \resizebox{0.98\linewidth}{!}{
    \begin{tabular}{l| r r| r r | r r r | r r r |c}
    \hline
      \textbf{Measure }& \multicolumn{2}{c}{\textbf{best single (C=1)}} &\multicolumn{2}{c}{ \textbf{best single + SVR}} & \multicolumn{3}{c}{\textbf{L-BFGS-B + SVR}} & \multicolumn{3}{c}{ \textbf{Multiple Paths + SVR}}  &\textbf{Limit~\cite{gordon1989anthropometric}}\\
      ~& mae [mm] & \% & mae [mm]  & \% & C & mae [mm] & \% & C & mae [mm] & \% &  [mm]\\
      \hline
      {\textbf{Male}} & \multicolumn{11}{c}{~} \\
  Ankle Circ.                             &36.4  & 0                  &8.4   &{\bf{37.6\%}}        &6  &12.3  &27.5\%     & 6   &{\bf{7.7}}  &28.6\%                    & 4 \\
  Bicep Circ.                            &7.8    &45.8\%         &6.5   &{\bf{57.8\%}}         &8  &19.4  &16.2\%     & 8   &{\bf{6.1}}  &57.6\%                    & 6 \\
  Calf Circ.                              &6.9    & 41.2\%         &4.0   &70.3\%                   &6  &17.7  &16.8\%     & 6  &{\bf{3.0}}  &{\bf{82.2\%}}        & 5 \\
  Chest Circ.                          &15.6   &60.0\%         &15.5  &61.6\%                   &5  &43.3  &22.9\%     & 5  &{\bf{14.3}} & {\bf{63.7\%}}        & 15 \\
  Elbow Circ.                           &4.0    & 56.5\%        &3.7    &62.8\%                   &8  &11.4  &16.2\%     & 8  &{\bf{2.6}}  &{\bf{77.9\%}}          & 4 \\
  Hip Circ.                                &9.2    & 71.7\%         &9.3   &{\bf{75.4\%}}         &4  &32.6  &22.9\%     & 4  &{\bf{8.8}}   &73.3\%                   & 12 \\
  Knee Circ.                             &8.5    &28.3\%         &6.0    &44.0\%                  &6  &15.5  &12.9\%     & 6  &{\bf{5.1}}    &{\bf{46.6\%}}         & 4 \\
  NaturalWaist Circ.                &15.8  &49.0\%         &13.2  &57.3\%                   &4  &50.0  &15.6\%     &4   &{\bf{12.8}}  &{\bf{57.6\%}}         &12\\
  NeckBase Circ.                     &35.1  &4.2\%            &10.2  &61.0\%                   &3  &15.7  &43.6\%     & 3  &{\bf{8.0}}    & {\bf{72.6\%}}        & 11 \\
  Neck Circ.                             &3.0    &92.1\%          &3.0   &91.1\%                     &4  &16.3  &22.4\%     & 4  &{\bf{2.5}}    &{\bf{93.7\%}}         & 6 \\
  Thigh Circ.                            &10.6   &31.4\%         &10.5  &32.5\%                    &8  &27.9  &16.8\%     & 8  &{\bf{7.9}}    &{\bf{48.7\%}}         & 6 \\
  TrouserWaist Circ.               &25.5   &-                    &12.0  &-                             &3  &36.4  &-               & 3   &{\bf{9.1}}    &-                              & - \\
  Wrist Circ.                             &7.2     &43.2\%          &5.2   &57.8\%                    &6  &6.6  &49.2\%     & 6  &{\bf{4.5}}    &{\bf{67.2\%}}          & 5 \\
  Shoulder\_to\_Shoulder      &13.7    &-                    &13.8 &-                              &4  &18.0  &-              & 4  &{\bf{12.0}}   &-                              & - \\
  Shoulder\_to\_Wrist            &40.3    &-                    &14.7 &-                              &6  &27.3  &-              & 6  &{\bf{12.7}}   &-                              & - \\
  {\textbf{Avg.}}              &16.0    &43.6\%          &9.1   &59.1\%                     &~  &23.4  &23.6\%     & ~  &{\bf{7.8}}    &{\bf{64.1\%}}          &~  \\
  \hline
  {\textbf{Female}} & \multicolumn{11}{c}{~} \\
  Ankle Circ.                            &18.8   &14.4\%            &14.3  &23.0\%             &6  &17.7  &14.7\%       & 6  &{\bf{13.4}}  &{\bf{24.7\%}}          & 4 \\
  Bicep Circ.                            &19.7    &8.5\%             &7.9   &48.3\%              &8  &15.9  &25.4\%       & 8  &{\bf{4.9}}   &{\bf{73.9\%}}           & 6 \\
  Calf Circ.                               &7.3     &37.4\%            &3.8   &70.7\%              &6  &18.0  &19.8\%       & 6  &{\bf{3.0}}    &{\bf{82.8\%}}         & 5 \\
  Bust Circ.                              &17.3    &44.0\%           &15.2  &60.6\%             &3  &42.1  &19.2\%       & 3  &{\bf{12.0}}  &{\bf{71.4\%}}            & 15 \\
  Elbow Circ.                           &4.5      & 57.4\%           &4.5   &59.7\%              &6  &11.7  &22.0\%      & 6   &{\bf{3.4}}   &{\bf{70.5\%}}          & 4 \\
  Hip Circ.                               &18.7     &26.3\%            &8.9  &70.9\%               &4  &37.0  &21.5\%      & 4   &{\bf{8.9}}   &{\bf{71.4\%}}          & 12 \\
  Knee Circ.                             &9.9      &21.1\%             &6.9   &39.4\%              &6  &17.3  &22.0\%      & 6   &{\bf{5.9}}    &{\bf{41.1\%}}         & 4 \\
  NaturalWaist Circ.               &13.7      &55.7\%           &12.8  &56.3\%              &5  &41.0  &16.4\%      & 5   &{\bf{12.0}}  &{\bf{59.7\%}}        & 12 \\
  NeckBase Circ.                    &58.8     & 0.6\%            &10.6   &{\bf{63.6\%}}   &3  &13.0  &54.2\%      & 3   &{\bf{10.2}}   &62.5\%                  & 11 \\
  Neck Circ.                             &6.3       &67.1\%           &5.5    &74.0\%              &5  &13.4  &32.2\%      & 5    &{\bf{4.8}}   &{\bf{81.5\%}}         & 6 \\
  Thigh Circ.                            &10.1      & 35.8\%         &9.7    &39.2\%              &8  &29.9  &13.6\%       & 8   &{\bf{7.9}}    &{\bf{46.3\%}}        & 6 \\
  TrouserWaist Circ.              &15.6      & -                   &15.4  &-                        &3  &38.0  &-                & 3    &{\bf{14.8}}  &-                             & - \\
  Wrist Circ.                             &6.0        &49.1\%          &5.0    &59.4\%            &8  &6.9  &40.7\%       & 8    &{\bf{4.4}}   &{\bf{65.7\%}}          & 5 \\
  UnderBust Circ.                    &14.2      &69.5\%          &14.3  &69.5\%            &2  &34.3  &27.1\%       & 2   &{\bf{13.4}}   &{\bf{71.8\%}}          & 16 \\
  Shoulder\_to\_Shoulder       &26.5     & -                   &13.8  &-                       &4  &17.9  &-                 & 4   &{\bf{12.7}}   &-                             & - \\
  Shoulder\_to\_Wrist             &22.4      & -                  &16.8  &-                       &4  &25.5  &-                & 4   &{\bf{13.7}}   & -                             & - \\
  Bust\_to\_Bust                      &12.2      &46.0\%         &11.6   &54.6\%            &9  &15.7  &39.6\%        & 9   &{\bf{10.4}}   &{\bf{57.5\%}}         & 10 \\
  NeckSide\_to\_Wrist             &26.4     & -                  &16.8   &-                      &4  &25.8  &-                  & 4   &{\bf{16.0}}  &-                               & - \\
  NeckSide\_to\_Bust              &13.9      &30.9\%         &13.4   &35.4\%           &6  &17.7  &24.9\%        & 6    &{\bf{13.0}}  &{\bf{36.6\%}}         & 8 \\
  {\textbf{Avg.}}                &17.0      &37.6\%         &10.9   &55.0\%           &~  &23.1  &26.2\%        & ~    &{\bf{9.7}}   &{\bf{61.2\%}}           & ~ \\
    \end{tabular}}
\end{table*}

\begin{figure*}
  \begin{center}
    \includegraphics[width=0.8\linewidth]{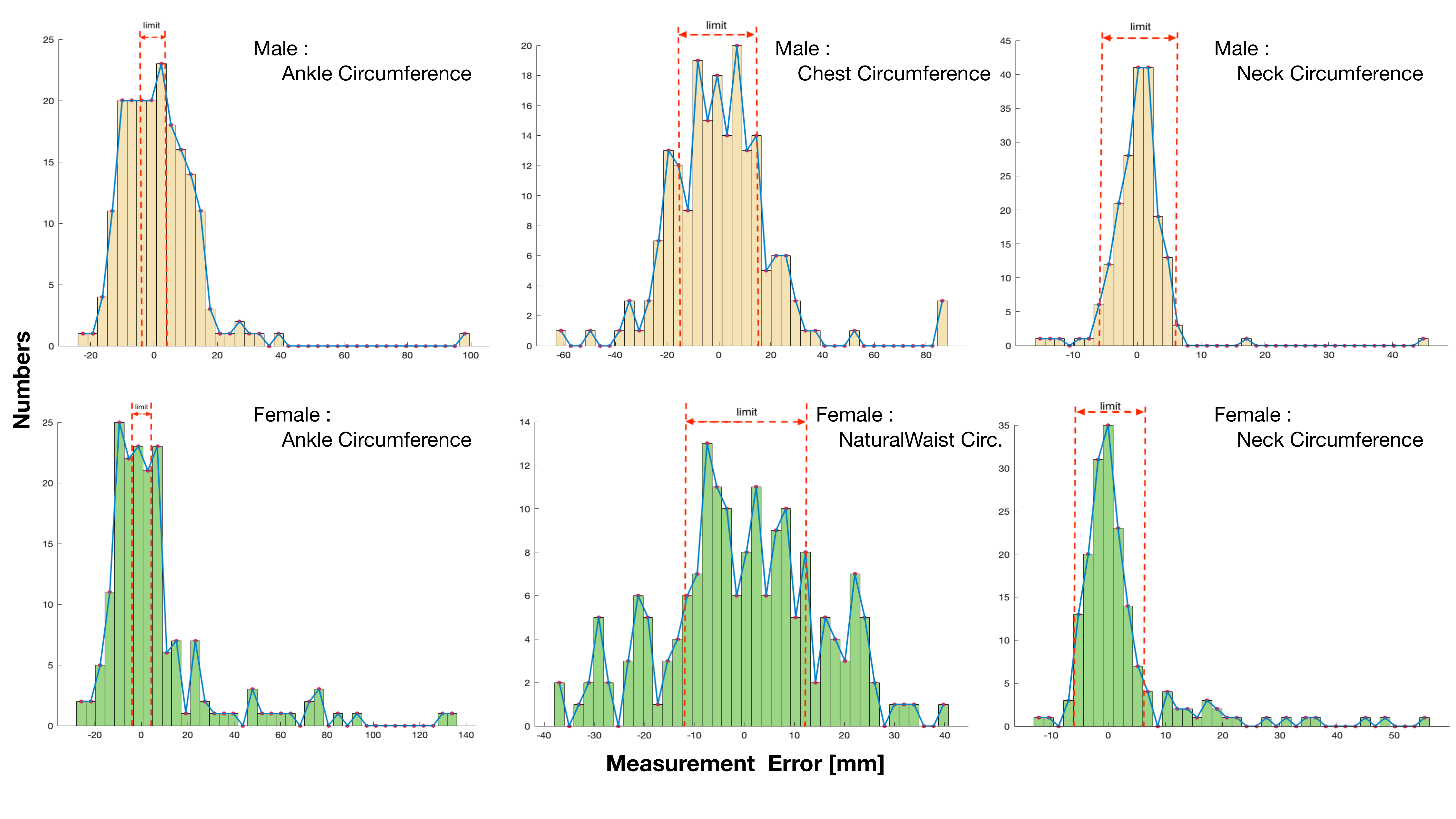}
    \caption{Error distributions illustrating low, moderate and well performing
      estimates. Top (Male) :
      {\em ankle circumference} (low),
      {\em chest circumference} (moderate),
      {\em neck circumference} (high); 
      Bottom (Female) :
      {\em ankle circumference} (low),
      {\em natural waist circumference} (moderate) and
      {\em neck circumference} (high). 
      The red vertical lines denote the acceptance thresholds in~\cite{gordon1989anthropometric}.
    \label{fig:distrs}}
    \end{center}
\end{figure*}

\begin{figure}[h]
  \centering
  \includegraphics[width=0.7\linewidth]{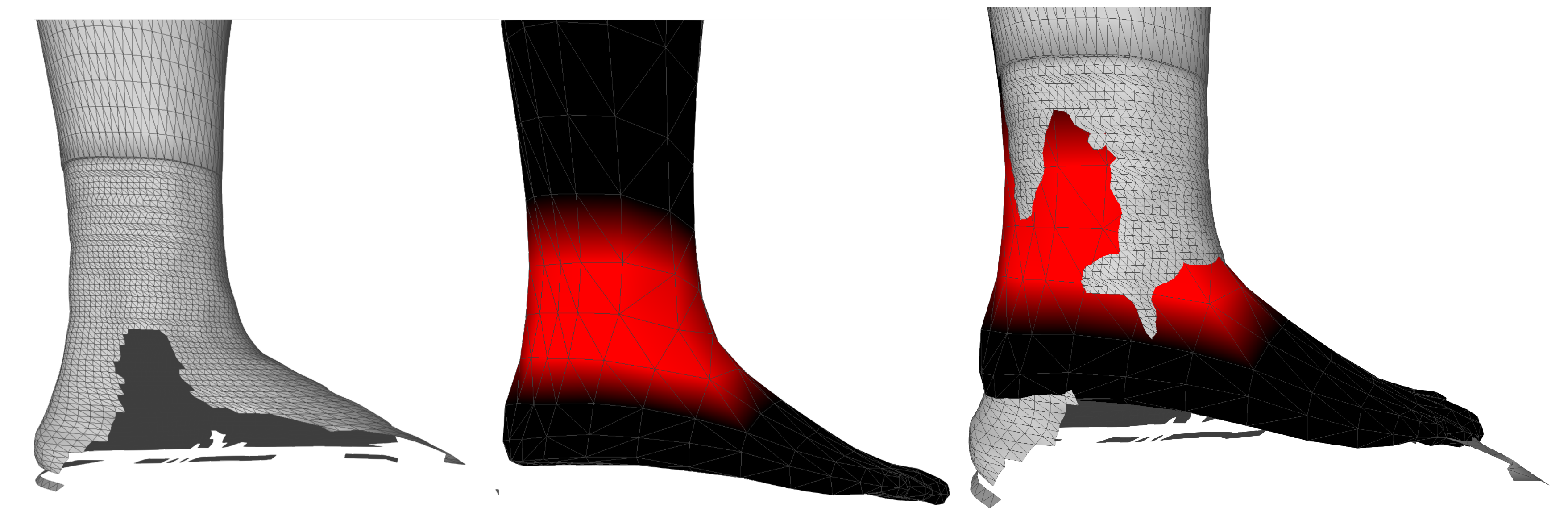}
  \includegraphics[width=0.7\linewidth]{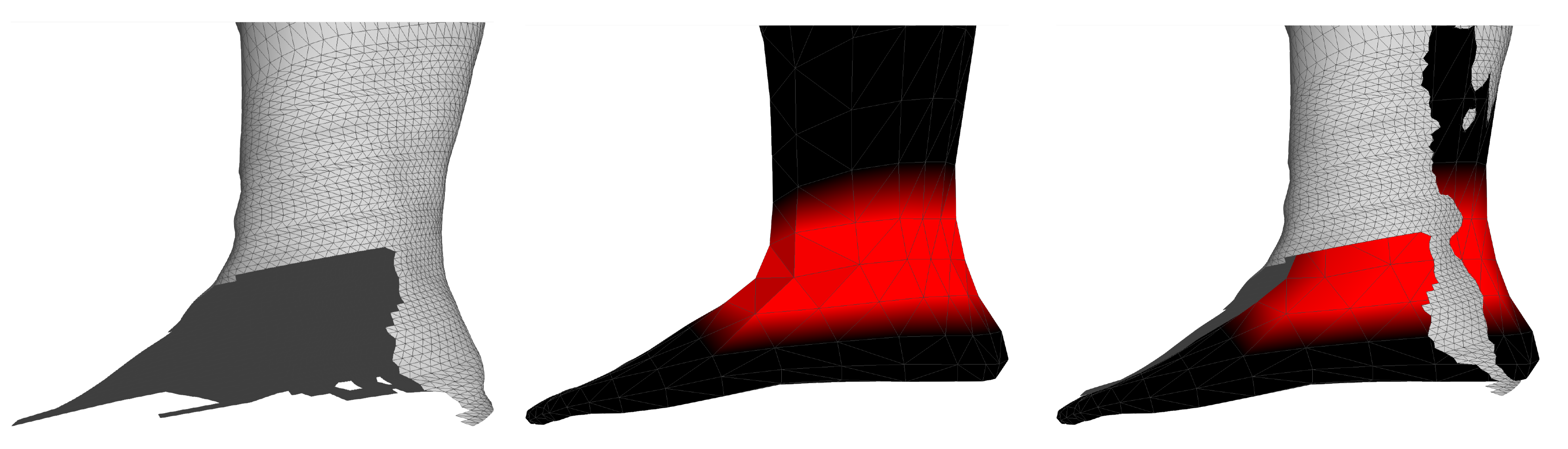}
  \caption{Two examples of registration failures due to missing points
    in the scanned point clouds:
    scanned point cloud (left),
    model emphasizing the ankle circumference location (middle) and
    output of the registration process (right) that illustrates the failure cases}
    \label{fig:ankle_case}
\end{figure}

The average 5-fold errors for each anthropometric measurement and their accuracy
thresholds and success rates are shown in Table~\ref{tab:results1}. In all cases,
the number of surface measurements were optimized for each anthropometric measurement
and the best performing regressor (non-linear SVR) was used. For the both male and female
subjects the best performing measurement was neck circumference with $93\%$ test cases below the
threshold ($6~mm$) for men and $81\%$ for women. The worst performing measure was
ankle circumference for which only $28\%$ of male
$24\%$ of female success rates were achieved. The error distributions for
  the male and female neck and ankle circumferences and male chest and female natural waist
  circumferences
are shown in Figure~\ref{fig:distrs}. The distributions reveal
that there exists a small amount of test samples with a large error. It turned
out that the main source of large estimation errors yields from the body scanner that often
misses certain body parts.
For example, feet regions often lack
point cloud points which makes the registration fail in these regions
(Figure~\ref{fig:ankle_case}).   

\subsection{Ablation Study}
\begin{figure}[!t]
  \centering
  \includegraphics[width=0.7\linewidth]{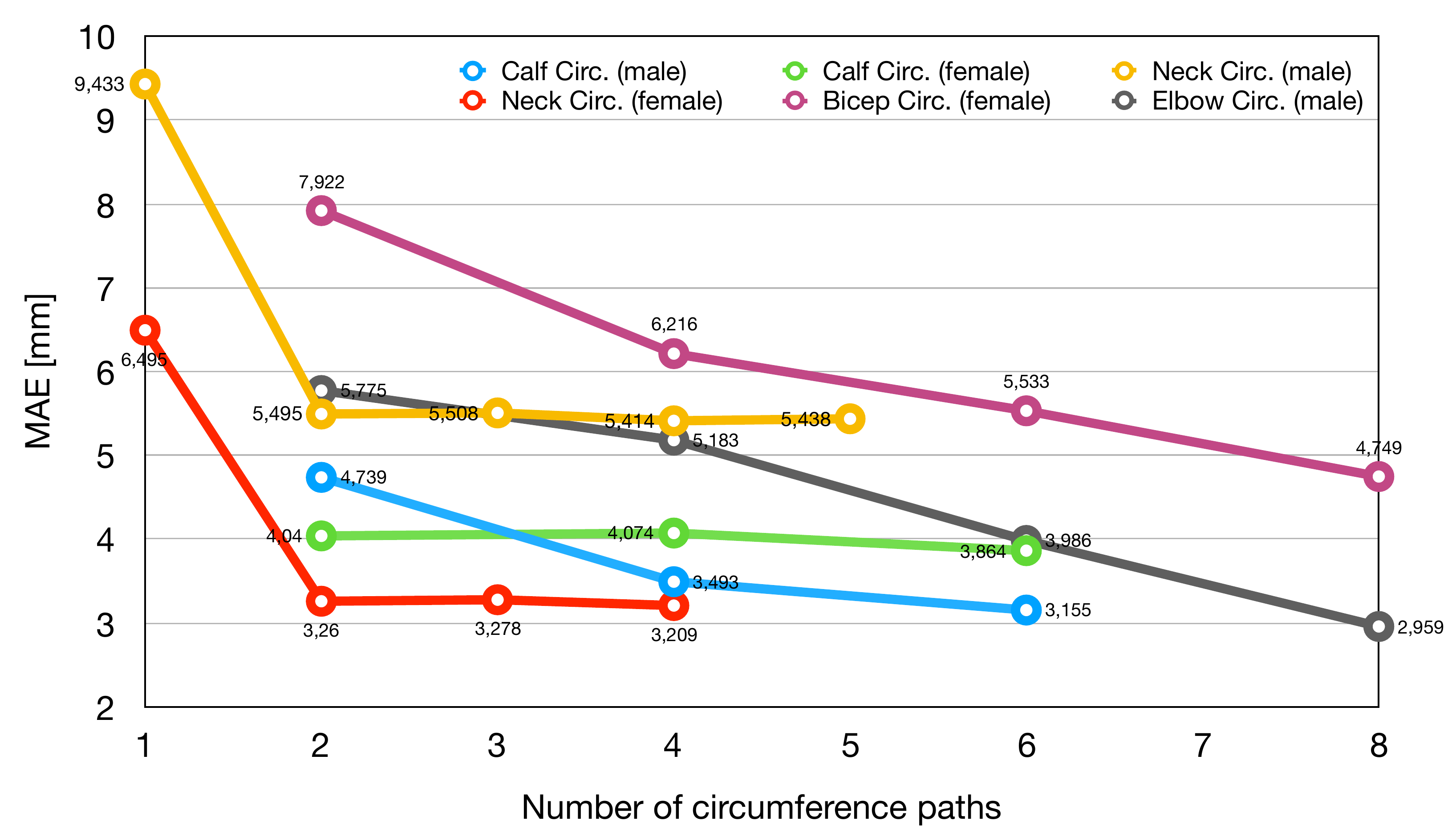}
  \includegraphics[width=0.7\linewidth]{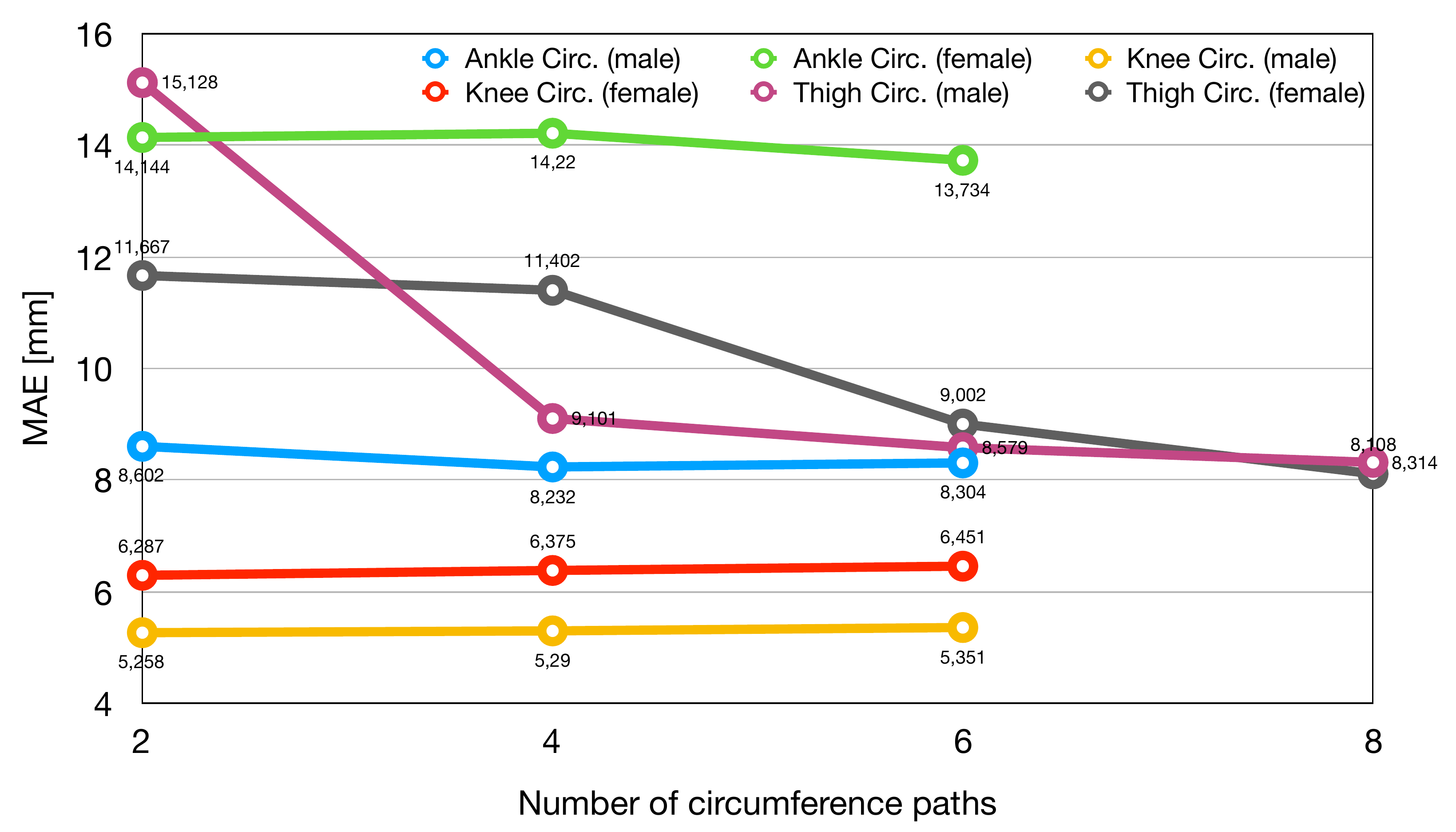}
  \caption{Test set errors (MAEs) as functions of the number of surface measurements (circumference paths) for
    three well performing (success $>50\%$) (top) and three poorly
    performing ($< 50\%$) anthropometric measurements (bottom)}
  \label{fig:circle_ablation_result}
\end{figure}

\paragraph{Number of circumference paths -- }
In the first ablation study, we investigated the effect of adding multiple surface
measurements (circumference paths) to the anthropometric regression. The results for three well and three poorly
performing measurements for the  both male and female are shown in Figure~\ref{fig:circle_ablation_result}.
Results are for {\textit{non-linear SVR}} regressor with 5-fold cross-validation. The most important
findings are that additional paths always improve the accuracy and depending on
the measurement the results saturate at $3$ to $9$ surface circumference paths. In particular,
paths close to the physical anthropometric measurement location strongly contribute to the
estimation accuracy.
The best single paths ($C=1$) were also selected using cross-validation and
    the results with and without SVR regression are shown in Table~\ref{tab:results1}. These
    results indicate that i) the multi-path regression is superior to single path regression
    and ii) SVR significantly improves the estimation performance.

\paragraph{Non-rigid ICP -- }
To validate the importance of non-rigid ICP we conducted an experiment where
the SMPL model was directly fitted to the point clouds. SMPL parameter optimization
was done using the popular L-BFGS-B optimizer~\cite{zhu1997algorithm}.
Similar to the non-rigid ICP, the {\textit{distance term}} $E_d$ with the normal
direction constraints was used as the target function. The stop criterion was set to
$10^{-6}$ to keep the computation times reasonable and the same pre-alignment procedure
was adopted. The results are shown in Table~\ref{tab:results1} and are clearly inferior
to the proposed non-rigid ICP registration.

\paragraph{Regression methods -- }
We compared a number of publicly available regression methods for the
regression step. The standard linear regressors were
{\textit{Linear Regression}},
{\textit{Stepwise Linear Regression}} and
{\textit{Ridge Regression}}, and more recent regression methods
are {\textit{Elastic Net Linear Regression}},
{\textit{Gaussian Process Regression (GPR)}},
{\textit{Binary Regression Decision Tree (BRDT)}},
{\textit{Linear Support Vector Regression (SVR)}} and
{\textit{Non-linear SVR}}. The mean accuracy and
success rates for these methods are shown in Table \ref{tab:results2}.
The results show that even the basic linear regressors (linear regression,
ridge regression and step-wise linear regression) perform well indicating
that the proposed registration step performs well.
Non-linear SVR and Gaussian Process Regression also perform well. They
are all safe choices for regressing anthropometric measurements from
surface measurements, but we selected the non-linear SVR due to its best
overall performance.
\begin{table*}[h]
  \caption{Average MAEs and success rates of several regression methods}
  \label{tab:results2}
  \begin{center}
    \resizebox{1.0\linewidth}{!}{
      \begin{tabular}{rrrrrrrrrrrrrrrr}
        \toprule
      {\em } & \multicolumn{15}{c}{{\em Error (MAE) [mm]}}\\
             \multicolumn{2}{c}{non-lin SVR}  & \multicolumn{2}{c}{Ridge Reg.}   &\multicolumn{2}{c}{Lin Reg.} &\multicolumn{2}{c}{Stepw. Reg.} & \multicolumn{2}{c}{GPR}  & \multicolumn{2}{c}{ElasticNet} & \multicolumn{2}{c}{BRDT} & \multicolumn{2}{c}{Lin SVR} \\
      \midrule
      \multicolumn{1}{c}{{\em Male}}\\
{\bf 7.81} & {\bf 64\%}   & 8.28    &  62\%            & 8.32                      & 62\%          & 8.35                    &  62\%       & 8.33                  & 62\%      & 9.16     & 56\%         & 10.94  & 52\%        & 31.80      & 19\%            \\
      \midrule
      \multicolumn{1}{c}{{\em Female}}\\
{\bf 9.73} & {\bf 61\%}  & 10.55                 &  57\%       & 10.62                  &  56\%          & 10.54    &  58\%           & 10.60                  &  59\%         & 12.05  &  50\%        & 14.05    &    46\%           & 28.72  & 23\%   \\
      \bottomrule
  \end{tabular}}
 \end{center}
\end{table*}

\section{Conclusions}
This work introduced a full processing pipeline for estimating physical anthropometric measurements
from 3D body scans. The pipeline consisted of a commercial 3D scanner,
a deformable SMPL body model, non-rigid ICP based model registration,
computation of circumference path features and non-linear regression for
anthropometric measurement estimation. Depending on the measurement our pipeline
provided success rates from $28\%$ to $93\%$ for male and from $24\%$ to $82\%$
for female subjects.
The proposed pipeline works in practice and shows that
an affordable scanning system can be built for clothing industry.

In the future work, we will further investigate and refine each step of the
pipeline. For example, selection of better surface features in addition to
the circumference paths, fast-to-compute alternatives for the slow
ICP algorithm (e.g. Chen et al.~\cite{chen2015robust}) and better scanners
and scanning procedures.

\bibliographystyle{spmpsci}      
\bibliography{human3d}

\end{document}